\documentclass[letterpaper]{article} 
\usepackage{aaai2026}  
\usepackage{times}  
\usepackage{helvet}  
\usepackage{courier}  
\usepackage[hyphens]{url}  
\usepackage{graphicx} 
\usepackage{natbib}  
\usepackage{caption} 
\usepackage{algorithm}
\usepackage{algorithmic}

\usepackage{algorithm}
\usepackage{algorithmic}
\usepackage{color}
\usepackage{pifont}

\newcommand{\ie}{\textit{i.e.}~}
\newcommand{\vs}{\textit{vs.}~}

\usepackage{amsmath,amssymb} 
\usepackage{color}

\usepackage{booktabs} 
\usepackage{amssymb}
\usepackage{multirow}

\usepackage{newfloat}
\usepackage{listings}
\DeclareCaptionStyle{ruled}{labelfont=normalfont,labelsep=colon,strut=off} 
\floatstyle{ruled}
\newfloat{listing}{tb}{lst}{}
\floatname{listing}{Listing}
\title{Causal-Tune: Mining Causal Factors from Vision Foundation Models for Domain Generalized Semantic Segmentation}
\author {
    Yin Zhang\textsuperscript{\rm 1,\rm 2}, 
    Yongqiang Zhang\textsuperscript{\rm 3}\thanks{Corresponding Author.},
    Yaoyue Zheng\textsuperscript{\rm 4,2}, 
    Bogdan Raducanu\textsuperscript{\rm 2}, 
    Dan Liu\textsuperscript{\rm 1}
}
\affiliations {
    \textsuperscript{\rm 1}School of Instrument Science and Engineering, Harbin Institute of Technology, Harbin, China\\
    \textsuperscript{\rm 2}Computer Vision Center, Universitat Autònoma de Barcelona, Barcelona, Spain\\  
    \textsuperscript{\rm 3}College of Computer Science, Inner Mongolia University, Inner Mongolia, China\\
    \textsuperscript{\rm 4}Institute of Artificial Intelligence and Robotics, Xi’an Jiaotong University, Shaanxi, China\\

    \{yin.zhang.hit, yaoyz105\}@gmail.com, 
    zhangyongqiang@imu.edu.cn,
    bogdan@cvc.uab.es, liudan@hit.edu.cn
}

\usepackage{bibentry}

\begin{document}

\maketitle

\begin{abstract}
Fine-tuning Vision Foundation Models (VFMs) with a small number of parameters has shown remarkable performance in Domain Generalized Semantic Segmentation (DGSS). Most existing works either train lightweight adapters or refine intermediate features to achieve better generalization on unseen domains.
However, they both overlook the fact that long-term pre-trained VFMs often exhibit artifacts, which hinder the utilization of valuable representations and ultimately degrade DGSS performance.
Inspired by causal mechanisms, we observe that these artifacts are associated with non-causal factors, which usually reside in the low- and high-frequency components of the VFM spectrum. In this paper, we explicitly examine the causal and non-causal factors of features within VFMs for DGSS, and propose a simple yet effective method to identify and disentangle them, enabling more robust domain generalization.
Specifically, we propose Causal-Tune, a novel fine-tuning strategy designed to extract causal factors and suppress non-causal ones from the features of VFMs. First, we extract the frequency spectrum of features from each layer using the Discrete Cosine Transform (DCT). A Gaussian band-pass filter is then applied to separate the spectrum into causal and non-causal components. To further refine the causal components, we introduce a set of causal-aware learnable tokens that operate in the frequency domain, while the non-causal components are discarded. Finally, refined features are transformed back into the spatial domain via inverse DCT and passed to the next layer. Extensive experiments conducted on various cross-domain tasks demonstrate the effectiveness of Causal-Tune. In particular, our method achieves superior performance under adverse weather conditions, improving +4.8\% mIoU over the baseline in snow conditions.  
\end{abstract}


\begin{links}
    \link{Code}{https://github.com/zhangyin1996/Causal-Tune}
\end{links}

\section{Introduction}
Recently, Vision Foundation Models (VFMs), such as DINOv2~\cite{dinov2}, CLIP~\cite{clip}, EVA02~\cite{eva, eva02}, and SAM~\cite{SAM}, have shown strong performance on various computer vision tasks. Specifically, using VFMs as backbones and fine-tuning them for Domain Generalized Semantic Segmentation (DGSS)~\cite{rein,fishertune,mamba_as} has led to significant improvements, even surpassing traditional CNN-based methods trained from scratch~\cite{resnet, shufflenetV2, zheng2025enhancing}.

\begin{figure}[t]
\centering
\includegraphics[width=\linewidth]{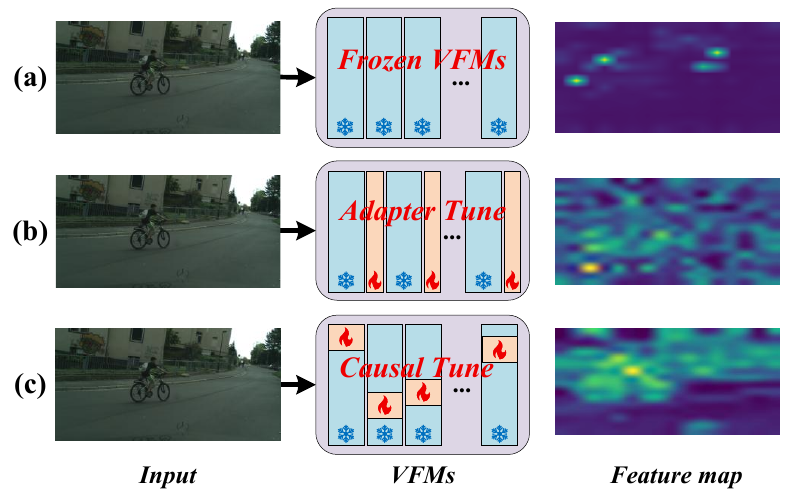}
\caption{
Visualization of DINOv2 feature maps. (a) Features extracted from the frozen DINOv2 contain noticeable artifacts. (b) Artifacts persist after applying existing adapter-based fine-tuning methods. (c) Our proposed Causal-Tune effectively suppresses these artifacts and guides the model to focus on domain-invariant causal factors. 
}
\label{fig_1}
\end{figure}

\begin{figure*}[t]
\centering
\includegraphics[width=\textwidth]{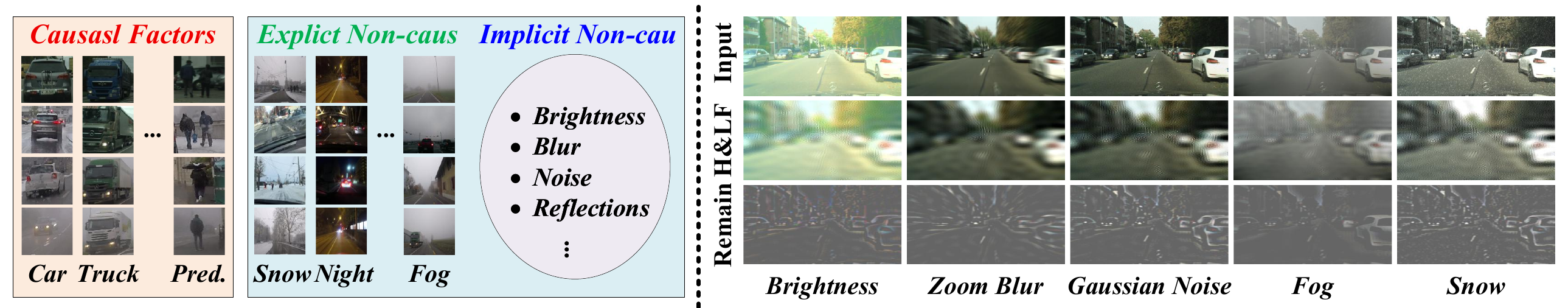}
\caption{Left: Causal factors and non-causal factors (contain explicit and implicit non-causal factors). Right: Visualization of images adding various non-causal factors actively after DCT, high- and low-frequency filtering (H\&LF), and inverse DCT.}
\label{fig_Causal_Ncau_Factors}
\end{figure*}

The predominant approach for fine-tuning VFMs is Parameter-Efficient Fine-Tuning (PEFT)~\cite{PEFT_survey}, where VFMs are kept frozen and only lightweight adapters are fine-tuned. 
Despite their superior performance compared to CNN-based models, transformer-based VFMs are often found to exhibit artifacts in extracted features~\cite{registers, vggt}. As illustrated in Figure~\ref{fig_1}(a), feature extraction with frozen DINOv2 reveals such artifacts. This phenomenon primarily arises from long-term pretraining on large-scale datasets. Although such training makes them highly expressive, it also introduces considerable redundancy in the extracted features, \ie, artifacts. Notably, as shown in Figure~\ref{fig_1}(b), adapter-based fine-tuning methods can improve feature map quality, yet artifacts still persist. Based on these observations, we hypothesize that these PEFT methods tend to indiscriminately fine-tune features from layers that may contain such artifacts. As a result, redundant features are not effectively suppressed, hindering the utilization of valuable representations from the VFM and ultimately degrading DGSS performance.

Recent advances in causal theory suggest that domain generalization can benefit from eliminating non-causal factors and mining the invariance of causal ones, as only causal factors tend to remain stable across domains~\cite{causality_cvpr2022oral, causality_pami, MAD, UFR, CWNet_LLIE}.
However, existing efforts in DGSS mainly focus on removing \textit{explicit} non-causal factors, like style-related information. They often overlook the fact that \textit{implicit} non-causal factors may also adversely affect performance~\cite{MAD}.
As illustrated on the left of Figure~\ref{fig_Causal_Ncau_Factors}, commonly used DGSS datasets contain two types of non-causal factors: (i) explicit ones, such as rain, snow, fog, and night; and (ii) implicit ones, such as brightness, blur, noise, and reflections. 
Prior works have largely ignored these implicit factors, which may help explain the remaining artifacts observed in Figure~\ref{fig_1}(b).
These observations naturally motivate our idea of removing both types of non-causal factors to purify VFM features and enhance generalization.

In this paper, we investigate two key questions and propose a method to explicitly identify and disentangle causal and non-causal factors in DGSS, to improve our understanding and enhance the performance on DGSS tasks.
\textbf{\textit{(1) How can causal and non-causal factors be effectively identified during the fine-tuning of VFMs?}} Frequency-based methods have long been considered effective for domain generalization. 
However, most works adopt the Fast Fourier Transform (FFT)~\cite{nussbaumer1981fast} or Haar Wavelet Transform (HWT)~\cite{HWT}, which often fail to identify implicitly non-causal factors and thus lead to suboptimal generalization performance. From a causal perspective, the frequency spectrum derived from Discrete Cosine Transform (DCT)~\cite{ahmed2006discrete} has been shown to better separate causal and non-causal factors~\cite{MAD}. Therefore, we alternatively utilize DCT to convert features from the spatial domain to the frequency domain before fine-tuning, to explicitly identify both causal and non-causal factors.
\textbf{\textit{(2) How can causal and non-causal factors be disentangled within the frequency domain?}} We actively inject various non-causal factors into images and then apply DCT, high and low frequency filtering (H\&LF), and inverse DCT to transform images from the frequency domain back to the spatial domain. The results are visualized on the right side of Figure~\ref{fig_Causal_Ncau_Factors}. We observe that non-causal factors tend to concentrate in both the high- and low-frequency components (second row). In contrast, residual components (last row) preserve structural and textural patterns that are relatively invariant across domains and are thus likely to represent causal factors.

To address the above challenges, we propose a simple but effective method: \textbf{Causal-Tune}. First, we extract the frequency spectrum of features from each layer using the DCT. A Gaussian band-pass filter is then applied to decompose the spectrum into causal and non-causal components. Then, the non-causal components are discarded, and only the causal components are retained for fine-tuning. A set of causal-aware learnable tokens is introduced to refine the causal components in the frequency domain. Finally, we use iDCT to transform the refined features back to the spatial domain and pass them to the next layer.
In summary, the main contributions of this paper are as follows:

\begin{itemize}
    \item 
    We explore Parameter-Efficient Fine-Tuning of VFMs for domain generalization semantic segmentation tasks from a causal perspective.
    
    \item We propose Causal-Tune, a novel fine-tuning strategy to mine causal factors and remove non-causal factors from the features of VFMs by a band-pass filter and then we utilize a set of learnable tokens to refine the causal factors from the frequency domain.

    \item Extensive experiments conducted on various cross-domain tasks show the effectiveness of Causal-Tune. In particular, Causal-Tune achieves superior performance in adverse weather conditions, improving +4.8\% mIoU over the baseline in `Snow' condition.
\end{itemize}

\section{Related Work}

\subsection{Domain Generalized Semantic Segmentation with Parameter-Efficient Fine-Tuning}

The goal of Domain Generalized Semantic Segmentation (DGSS) is to enhance a model’s generalization ability on unseen domains by training only in a source domain. Recently, VFMs have demonstrated strong representation abilities thanks to large-scale pretraining, and DGSS methods based on Parameter-Efficient Fine-Tuning (PEFT) have attracted increasing attention. Rein~\cite{rein} is the first work to leverage PEFT for DGSS. They propose an embedded fine-tuning mechanism that inserts a small number of trainable parameters to refine feature maps between layers within the VFM. Subsequent PEFT-based DGSS works, such as SET~\cite{SET}, design spectrally decomposed tokens to learn style-invariant features, while FADA~\cite{FADA} proposes a frequency-adapted learning scheme to decouple style information from domain-invariant content. However, both of the above methods overlook an important fact: long-term pretrained VFMs often exhibit artifacts, and indiscriminately fine-tuning them can negatively impact performance on downstream tasks. We observe that artifacts are often associated with non-causal factors and aims to investigate this issue from a causal perspective.

\subsection{Frequency-based Domain Generalization}

Many works applying frequency domain transformation methods for domain generalization have demonstrated that frequency-based information is robust to domain shifts. For instance, FDA~\cite{FDA} utilizes the FFT to swap the amplitude spectrum between images, achieving strong performance in DGSS. FACT~\cite{FACT} addresses domain generalization by mixing the amplitude spectra of image pairs from different source domains. More recently, fine-tuning VFMs in the frequency domain for DGSS has attracted growing interest. SET~\cite{SET} transforms the features of VFMs into the frequency domain using FFT, and learns style-invariant features through a set of learnable tokens. Similarly, FADA~\cite{FADA} leverages the HWT to extract style information from the frequency domain. However, they overlook the fact that long-term pre-trained VFMs often exhibit artifacts, which hinder the utilization of valuable representations and ultimately degrade DGSS performance. Different from previous works, we explore the relationship between the frequency domain and causal factors in DGSS, aiming to enhance generalization performance by removing non-causal factors.

\subsection{Causal Mechanism in Domain Generalization}

Causal mechanism suggests that domain generalization can benefit from mining invariant causal factors while removing non-causal factors, inspiring new methodologies for improving the generalization ability of models. Specifically, ~\cite{causality_cvpr2022oral} proposes a representation learning method to extract causal factors from inputs, thereby boosting generalization performance. ~\cite{causality_pami} introduces a causal intervention reasoning module to learn invariant feature representations for multiple adverse weather generalization. UFR~\cite{UFR} constructs a structural causal model to analyze the limitations of domain generalization and designs a causal attention learning module to address these challenges. The most similar work to ours is MAD~\cite{MAD}, which removes implicit non-causal factors by a multi-view adversarial discriminator. However, the main difference between MAD and our work is that MAD relies on a data-augmentation strategy, whereas our method directly fine-tunes the features of VFMs after the explicit disentangle of causal and non-causal factors.

\begin{figure*}[t]
\centering
\includegraphics[width=0.98\textwidth]{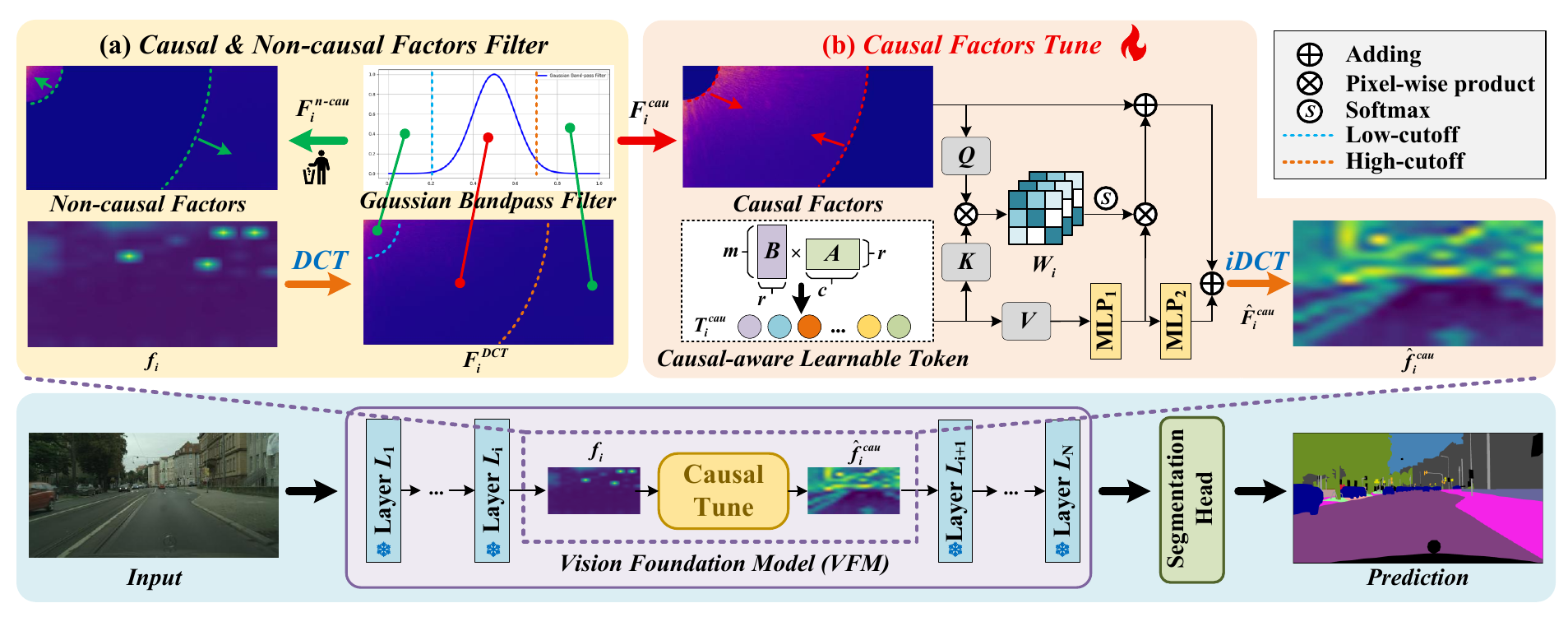}
\caption{The pipeline of our proposed Causal-Tune. (a) The output feature $f_{i}$ of layer $L_{i}$ are first transformed to the frequency domain feature $F^{DCT}_i$ using DCT, and then a Gaussian band-pass filter separates it into causal factors $F^{cau}_i$ (red) and non-causal factors $F^{n-cau}_i$ (green). Only the causal factors are used for subsequent fine-tuning, while the non-causal factors are discarded. (b) A series of causal-aware learnable tokens $T^{cau}_{i}$ interacts with the causal factors $F^{cau}_i$ through an attention mechanism to refine them. The refined causal factors $\hat{F}^{cau}_i$ are then transformed back to the spatial domain $\hat{f}^{cau}_i$ using the iDCT.}
\label{fig_pipeline}
\end{figure*}

\section{Proposed Method}

\subsection{Preliminaries}

Given a frozen VFM with $N$ layers ($L_1, L_2, ..., L_N$), the output feature of the $i$-th layer $L_i$ is denoted as $f_i$. The feature propagation from layer $L_i$ to $L_{i+1}$ can be expressed as:
\begin{equation}
f_{i+1} = L_{i+1}(f_{i}).
\end{equation}
The core idea of fine-tuning VFMs in DGSS is to refine the features between layers to improve the generalization on unseen domains, which can be formulated as:
\begin{equation}
\hat{f}_{i+1} = L_{i+1}(f_{i} + \Delta f_{i}).
\end{equation}
As $f_i$ remains frozen, the goal is to propose a method to generate refined features $\Delta f_i$. To this end, we design a novel fine-tuning method, which formulates the refinement as $\Delta f_{i} = \text{CausalTune}(f_{i})$. The proposed method leverages causal mechanism to enhance feature representations by identifying and preserving causal factors while filtering out non-causal ones.

\subsection{Causal \& Non-causal Factors Filter}

As discussed above, the spectrum obtained by the Discrete Cosine Transform (DCT) provides a more effective representation for distinguishing between causal and non-causal factors. In particular, the extremely high- and low-frequency components of the spectrum are more likely to encode non-causal information, while the remaining parts tend to capture more causal factors. Based on this observation, we adopt a simple yet effective approach by applying a band-pass filter to separate causal and non-causal components. Specifically, as illustrated in Figure~\ref{fig_pipeline}(a), for the output features $f_{i}(h,w)$ extracted from layer $L_i$ of the frozen VFM, we first use the DCT to transform $f_{i}(h,w)$ into the frequency domain. This process is mathematically formulated as:

\begin{multline}
F^{DCT}_i(u,v) = \alpha(u)\alpha(v) \sum_{h=0}^{H-1} \sum_{w=0}^{W-1} f_{i}(h,w) \cdot \\
\cos\left( \frac{\pi (2h+1)u}{2H} \right) \cdot
\cos\left( \frac{\pi (2w+1)v}{2W} \right) ,\\
\alpha(u) = 
\begin{cases}
\sqrt{\frac{1}{H}},  u = 0 \\
\sqrt{\frac{2}{H}},  u > 0
\end{cases}
\alpha(v) = 
\begin{cases}
\sqrt{\frac{1}{W}},  v = 0 \\
\sqrt{\frac{2}{W}},  v > 0
\end{cases}
\end{multline}
where  $F^{DCT}_i(u,v)$ denotes the frequency domain features after DCT. $H$ and $W$ are the height and width of the features. $\alpha(u)$ and $\alpha(v)$ are the normalization factors that ensure the energy conservation of DCT and inverse Discrete Cosine Transform (iDCT).

Then, we apply a Gaussian band-pass filter to separate causal and non-causal factors (features):
\begin{equation}
G(u,v) = \exp\left(-\frac{u^2 + v^2}{2R_H^2}\right) 
- \exp\left(-\frac{u^2 + v^2}{2R_L^2}\right),
\end{equation}
where $R_{H}$ and $R_{L}$ denote the cutoff frequency for high- and low-frequency, respectively. As a result, the causal features $F^{cau}_i$ and non-causal features $F^{n-cau}_i$ can be obtained from the frequency domain by using a Gaussian band-pass filter:
\begin{equation}
\begin{aligned}
F^{cau}_i &= F^{DCT}_i(u,v) \cdot G(u,v), \\
F^{n-cau}_i &= F^{DCT}_i(u,v) \cdot (1 - G(u,v)).
\end{aligned}
\end{equation}
As shown in Figure~\ref{fig_pipeline}(a), the components below the low cut-off frequency $R_{L}$ and above the high cut-off frequency $R_{H}$ are considered as non-causal factors $F^{n-cau}_i$ (green), while the remaining components are considered as causal factors $F^{cau}_i$ (red). The objective of our method is to mine causal factors from the VFMs, as they contain domain-invariant information, whereas non-causal factors primarily encode domain-specific information. Thus, the non-causal factors $F^{n-cau}_i$ are discarded, and only the causal factors $F^{cau}_i$ are retained for fine-tuning.

\begin{table*}[t!]
\small
\centering
\begin{tabular}{l| c c c c c| c c |c c c}
\toprule
 &\multicolumn{5}{c|}{Trained on $C.$ $\rightarrow$ ACDC}  &\multicolumn{2}{c|}{Trained on $C.$} &\multicolumn{3}{c}{Trained on $G.$}\\
\midrule
Method &Night &Snow &Fog &Rain &Avg. &$\rightarrow B.$ &$\rightarrow M.$ &$\rightarrow C.$ &$\rightarrow B.$ &$\rightarrow M.$\\
\midrule
\textit{Resnet-based}: \\
IBN~\cite{IBN} &21.2 &49.6 &63.8 &50.4 &46.3   &48.56 &57.04 &33.85 &32.30 &37.75\\
Itenorm~\cite{Itenorm} &23.8 &49.9 &63.3 &50.1  &46.9  &49.23  &56.26  &31.81 &32.70 &33.88\\
IW~\cite{IW} &21.8 &47.6 &62.4 &52.4 &46.1  &48.49 &55.82 &29.91 &27.48 &29.71\\
ISW~\cite{ISW} &24.3 &49.8 &64.3 &56.0  &48.6  &50.73 &58.64 &36.58 &35.20 &40.33\\
DIRL~\cite{DIRL} &- &- &- &- &- &51.80 &- &41.04 &39.15 &41.60\\
\midrule
\textit{Transformer-based}: \\
HGFormer~\cite{HGFormer} &52.7 &68.6 &69.9 &72.0  &65.8 &53.40 &66.90 &- &- &-\\
CMFormer~\cite{CMFormer}  &33.7 &64.3 &77.8 &67.6  &60.9  &59.27 &71.10 &55.31 &49.91 &60.09\\
\midrule
\textit{VFM-based}: \\
Rein~\cite{rein}  &55.9 &70.6 &79.5 &72.5  &69.6  &63.54 &74.03 &66.40 &60.40 &66.10\\
SET~\cite{SET} &\underline{57.3} &\underline{73.6} &80.1 &74.8  &\underline{71.5}  &65.07 &75.67 &\underline{68.06} &61.64 &67.68 \\
FADA~\cite{FADA} &\textbf{57.4} &73.5 &\underline{80.2} &\underline{75.0}  &\underline{71.5}   &\underline{65.12} &\underline{75.86} &\textbf{68.23} &\textbf{61.94} &\underline{68.09}  \\
Ours &56.2 &\textbf{75.4} &\textbf{81.3} &\textbf{75.2}  &\textbf{72.0}  &\textbf{66.28} &\textbf{76.05} &66.22 &\underline{61.80} &\textbf{68.21}  \\
\midrule
\midrule
\textit{Comparison with Baseline} (Rein):   &+0.3 &+4.8 &+1.8 &+2.7 &+2.4   &+2.74 &+2.02 &-0.18 &+1.40 &+2.11  \\
\bottomrule
\end{tabular}
\caption{DGSS performance of our proposed Causal-Tune. $G.$, $C.$, $B.$ and $M.$ denotes GTA5, Cityscapes, BDD100K and Mapillary datasets, respectively. `-' indicates that the results are not reported in the original paper and no source code is available. Note that in order to be consistent with previous methods, only report one decimal official results under $C. \rightarrow$ ACDC setting. \textbf{Blod} and \underline{underline} results represent best and second-best performance, respectively.
}
\label{table_results}
\end{table*}

\subsection{Causal Factors Tune}

Building on awareness of domain-invariant causal factors, we introduce a learnable token $T^{cau}_{i}$ to fine-tune these causal factors within each VFM layer. As shown in Figure~\ref{fig_pipeline}(b), the learnable token of layer $L_{i}$ can be represented as:
\begin{equation}
T^{cau}_{i}  = B_i  A_i, \ \ \ \   T^{cau}_{i}\in \mathbb{R}^{m \times c},
\end{equation}
where $B_i \in \mathbb{R}^{m \times r}$ and $A_i \in \mathbb{R}^{r \times c}$, with $m$ representing the length of the sequence $T^{cau}_i$, and $c$ denoting the dimensionality of the feature $f_i$.

To allow each learnable token to capture task-specific knowledge and improve segmentation performance, we utilize an attention-based mechanism to enhance the semantic clarity of causal factors~\cite{tqdm, SET}. Specifically, we consider causal factors $F^{cau}_i$ as \textit{Query}, causal-aware learnable token $T^{cau}_{i}$ as \textit{Key} and \textit{Value}.
The attention weights $W_i$ are computed through a pixel-wise product between $F^{cau}_i$ and $T^{cau}_i$.
Then, the token $T^{cau}_i$ (\textit{Value}) are projected into the feature space of $F^{cau}_i$ using a multi-layer perceptron $MLP_1$. Additionally, a residual connection is incorporated to better align the token $T^{cau}_i$ with the corresponding causal factors $F^{cau}_i$.
Finally, an additional MLP layer, denoted as $MLP_2$, is applied for flexible feature refinement. Residual connections are incorporated to mitigate vanishing gradient issues. The refined causal factors $\hat{F}^{cau}_i$ in the frequency domain are computed as follows:
\begin{equation}
W_{i} = \text{Softmax}\left(\frac{F^{cau}_{i} \times T^{cau \top}_{i}}{\sqrt{c}}\right),
\end{equation}
\begin{equation}
\tilde{F}^{cau}_i \leftarrow F^{cau}_i + W_i \times MLP_1(T^{cau}_i),
\end{equation}
\begin{equation}
\hat{F}^{cau}_i = F^{cau}_i + MLP_2(\tilde{F}^{cau}_i),
\end{equation}
where the softmax activation function is used to normalize the attention weights.
Furthermore, an inverse Discrete Cosine Transform (iDCT) is required to convert the refined causal features from the frequency domain back to the spatial domain. The detailed process is described as follows:
\begin{multline}
\hat{f}^{cau}_{i}(h,w) = \alpha(u) \alpha(v) \sum_{u=0}^{H-1} \sum_{v=0}^{W-1} \hat{F}^{cau}_i(u,v) \cdot \\
\cos\left( \frac{\pi(2h+1)u}{2H} \right) \cdot
\cos\left( \frac{\pi(2w+1)v}{2W} \right) \\
\alpha(u) =
\begin{cases}
\sqrt{\frac{1}{H}}, u = 0 \\
\sqrt{\frac{2}{H}}, u > 0
\end{cases}
\alpha(v) =
\begin{cases}
\sqrt{\frac{1}{W}}, v = 0 \\
\sqrt{\frac{2}{W}}, v > 0
\end{cases}
\end{multline}
where $\hat{f}^{cau}_{i}(h,w)$ denotes refined causal features in the spatial domain.

\section{Experiments}
\subsection{Benchmarks and Training Setting}
\subsubsection{Datasets and Metric.}
We evaluate the performance of Causal-Tune on five driving-scene datasets that share 19 categories for DGSS. \textbf{Cityscapes}~\cite{cityscapes} is a widely used dataset for autonomous driving, which contains 2975 training images and 500 validation images with a resolution of 2048$\times$1024. \textbf{GTA5}~\cite{gta5} provides 24966 images with 1914$\times$1052 resolution obtained from the game engine.
\textbf{ACDC}~\cite{acdc} comprises 406 images for validation, which covers four adverse conditions, \ie, night, snow, fog, and rain. \textbf{BDD100K}~\cite{bdd100k} is a large-scale dataset that consists of 70k training images and 10k validation images. \textbf{Mapillary}~\cite{Mapillary} has 18000 and 2000 images for training and validation, respectively. Following previous works~\cite{rein,SET}, we adopt the mean Intersection of Union (mIoU) as the evaluation metric.

\subsubsection{Implementation details.}
Following previous VFM-based fine-tuning methods on DGSS, we choose DINOv2~\cite{dinov2} and Mask2Former~\cite{mask2former} as the default VFM and segmentation head. AdamW is usd as the optimizer with a base learning rate of 1e-4. We set $R_L$ = 0.2 and $R_H$ = 0.7 in Gaussian band-pass filter. The model is trained 40000 iterations with a batch size of 4 and all the resolution of input images is 512 $\times$ 512. Our method is implemented based on \textit{MMSegmentation}~\cite{mmseg} with a single RTX3090 GPU, occupying 14 GB memory.

\subsection{Comparison with SOTA Methods}
We use Rein~\cite{rein} as baseline and following by Rein, we compare our method with existing DGSS on three setups: i) Cityscapes ($C.$) $\rightarrow$ ACDC, ii) Cityscapes ($C.$) $\rightarrow$ BDD100K ($B.$), Mapillary ($M.$) and iii) GTA5 ($G.$) $\rightarrow$ Cityscapes ($C.$), BDD100K ($B.$), Mapillary ($M.$). In addition, we also compare with three types of DGSS methods: 1) ResNet-based, 2) Transformer-based, and 3) VFM-based methods, to verify the effectiveness of our Causal-Tune. All of our experiments are implemented and averaged by three independent repetitions.

As shown in Table~\ref{table_results}, Causal-Tune outperforms all Resnet-based and Transformer-based methods. In addition, by elaborately designing the causal factors tuning, our method achieves the best results in 7 out of 10 cases and the second-best results in 1 case compared to VFM-based methods. The detailed results are as follows:

\subsubsection{i) Cityscapes ($C.$) $\rightarrow$ ACDC.}
As illustrated in the second column of Table~\ref{table_results}, our method has an obvious advantage in performance owing to its Causal \& Non-causal Factors Filter and Causal Factor Tune strategy. It achieves the best results on `Snow', `Fog', and 'Rain' weather conditions, outperforming SET~\cite{SET} and FADA~\cite{FADA} by (1.8\% \& 1.9\%), (1.2\% \& 1.1\%), and (0.4\% \& 0.2\%) respectively. Furthermore, compared to baseline method Rein~\cite{rein} (in the last row of Table~\ref{table_results}), Causal-Tune brings a 
2.4\% improvement (72.0\% \vs 69.6\%) on average, where the most notable improvement is on `Snow' condition (from 70.6\% to 75.4\%, +4.8\%). Moreover, our method achieves a large margin improvement (from 72.5\%
to 75.2\%, +2.7\%) on `Rain' condition, an impressive improvement of +1.8\% is obtained on `Fog' condition, and a satisfactory result on `Night' condition (56.2\% \vs 55.9\%). The above comparison clearly demonstrates the effectiveness of our proposed method on adverse weather conditions.
\subsubsection{ii) Cityscapes ($C.$) $\rightarrow$ BDD100K ($B.$), Mapillary ($M.$).}
As illustrated in the third column of Table~\ref{table_results}, Causal-Tune achieves a new SOTA performance under $C.\rightarrow B.$ (66.28\%) and $C. \rightarrow M.$ (76.05\%) setting, respectively.

\subsubsection{iii) GTA5 ($G.$) $\rightarrow$ Cityscapes ($C.$), BDD100K ($B.$), Mapillary ($M.$).}

As shown in the fourth column of Table~\ref{table_results}, compared with the SOTA method FADA~\cite{FADA}, our method obtains the best result on $G. \rightarrow M.$ setting (68.21\%). Moreover, our method can still achieve competitive results under $G. \rightarrow B.$ (61.80\% \vs 61.94\%) setting. 

However, our method achieves a worse performance under $G. \rightarrow C.$ setting. We think the reason is that GTA5~\cite{gta5} is a synthetic dataset while Cityscapes~\cite{cityscapes} is obtained from real scenes. Our method may not be able to mine causal factors effectively under \textit{synthetic-to-real} generalization. 

In summary, our proposed method not only obtains an excellent performance in \textit{real-to-real} setting ($C.$ $\rightarrow$ ACDC and  $C.\rightarrow B.,M.$) but also maintains a satisfactory performance in \textit{synthetic-to-real} setting ($G.\rightarrow C.,B.,M.$).

\subsection{Ablation Studies and Analysis}

\begin{table}[t]
\small
\centering
\begin{tabular}{l| c c c c c}
\toprule
    &\multicolumn{5}{c}{Trained on $C.$ $\rightarrow$ ACDC} \\
\midrule
Method &Night &Snow &Fog &\multicolumn{1}{c|}{Rain} &Avg.\\
\midrule
FFT &\textbf{57.1} &71.1 &\underline{80.6} &\multicolumn{1}{c|}{69.2} &\underline{69.5}  \\
HWT &52.8 &\underline{72.2} &79.4 &\multicolumn{1}{c|}{\underline{72.4}} &69.2  \\
DCT (Ours) &\underline{56.2} &\textbf{75.4} &\textbf{81.3} &\multicolumn{1}{c|}{\textbf{75.2}}
&\textbf{72.0}  \\
\midrule
\midrule
    &\multicolumn{2}{c|}{Trained on $C.$} &\multicolumn{3}{c}{Trained on $G.$}\\
\midrule
Method &$\rightarrow B.$  &\multicolumn{1}{c|}{$\rightarrow M.$ } &$\rightarrow C.$  &$\rightarrow B.$  &$\rightarrow M.$ \\
\midrule
FFT &\underline{65.07} &\multicolumn{1}{c|}{71.00} &65.22 &60.56 &64.77  \\
HWT &64,43 &\multicolumn{1}{c|}{\underline{71.52}} &\underline{65.77} &\underline{61.08} &\underline{66.66} \\
DCT (Ours) &\textbf{66.28} &\multicolumn{1}{c|}{\textbf{76.05}} &\textbf{66.22} &\textbf{61.80} &\textbf{68.21}\\
\bottomrule
\end{tabular}
\caption{Analysis of different frequency domain
transformation methods. 
}
\label{table_frequency}
\end{table}

\begin{table}[t]
\small
\centering
\begin{tabular}{c c c| c c c c c}
\toprule
\multicolumn{3}{c|}{Frequency Filter} &\multicolumn{5}{c}{Trained on $C.$}\\
\midrule
\textit{LF} &\textit{HF} & \textit{L\&HF} &Night &Snow &Fog &Rain &Avg.\\
\midrule
- &- &- &55.9 &70.6 &79.5 &72.5 &69.6\\
\checkmark &- &- &\textbf{57.3} &72.1 &73.2 &71.1 &68.4\\
- &\checkmark &- &54.8 &\underline{73.7} &\underline{80.5} &\underline{74.1}  &\underline{70.4}\\
- &- &\checkmark &\underline{56.2} &\textbf{75.4} &\textbf{81.3} &\textbf{75.2} &\textbf{72.0}\\
\bottomrule
\end{tabular}
\caption{Ablation studies on different filtering methods for DCT frequency spectrum. \textit{LF} is low-frequency filtering, \textit{HF} is high-frequency filtering and \textit{L\&HF} denotes both low and high-frequency filtering, \ie, a band-pass filter.
}
\label{table_DCT_component}
\end{table}

\begin{figure}[t]
\centering
\includegraphics[width=\linewidth]{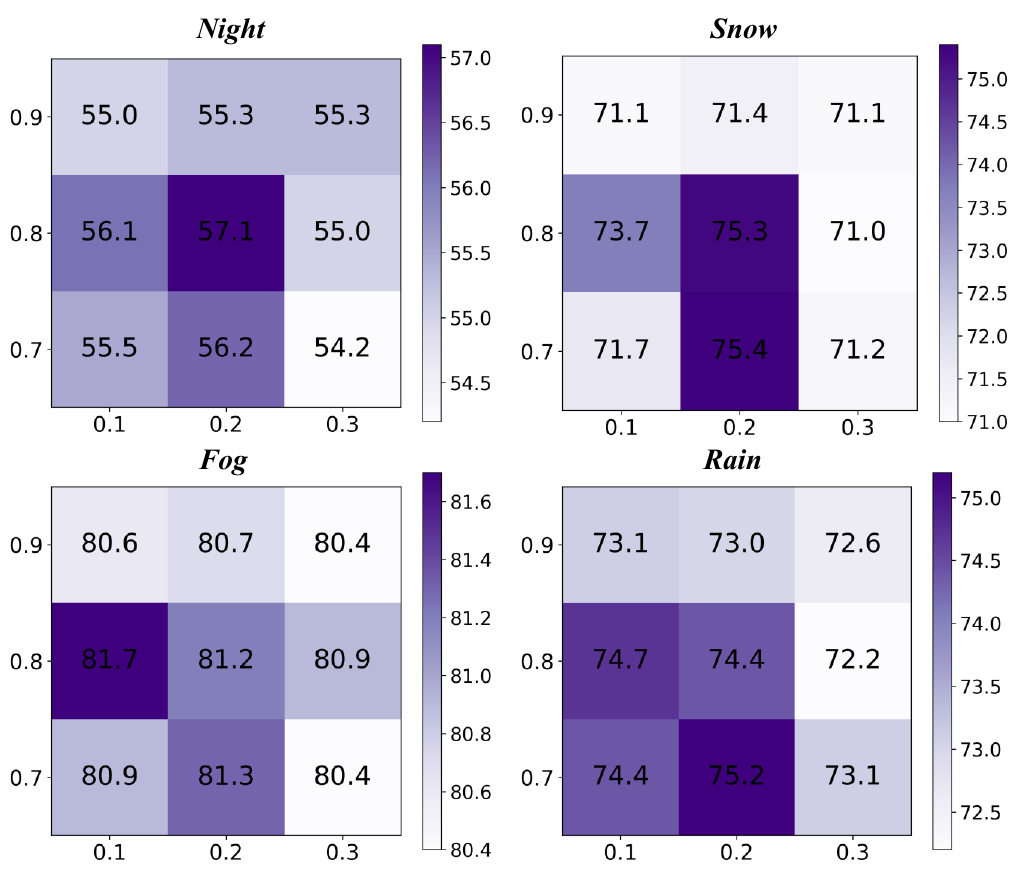}
\caption{Accuracy matrix of cutoff frequency analysis under $C. \rightarrow$ ACDC generalization. The horizontal and vertical axis are low- and high-cutoff frequency, respectively.}
\label{fig_acc_matrix}
\end{figure}

\begin{figure*}[t]
\centering
\includegraphics[width=0.95\textwidth]{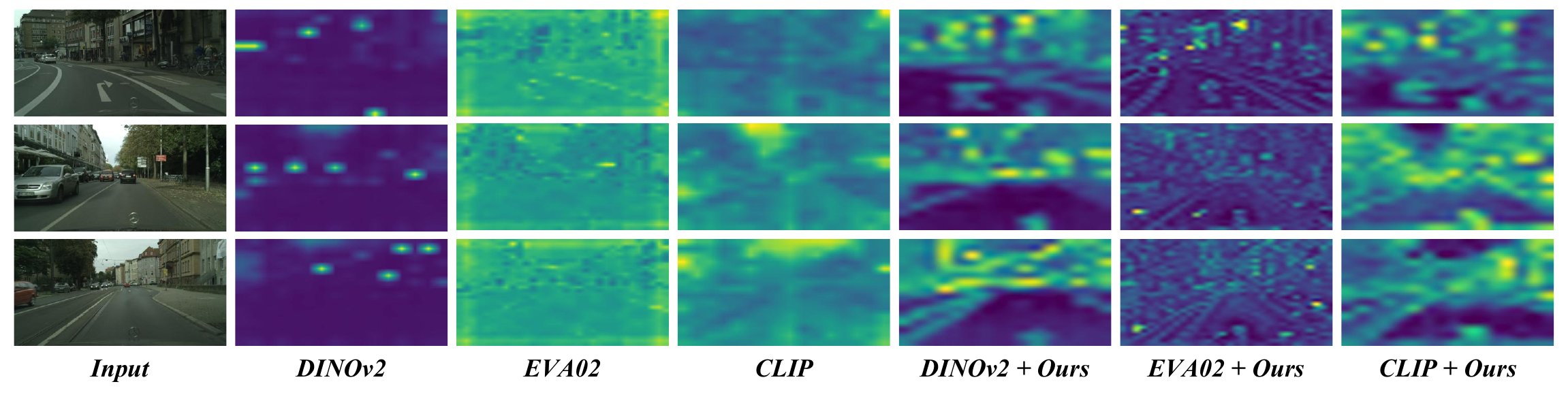}
\caption{of feature map from different VFMs. We show the feature maps of frozen DINOv2~\cite{dinov2}, EVA02~\cite{eva02}, CLIP~\cite{clip} and after fine-tuning by our method.}
\label{fig_feature_map}
\end{figure*}

\begin{figure*}[t]
\centering
\includegraphics[width=0.98\textwidth]{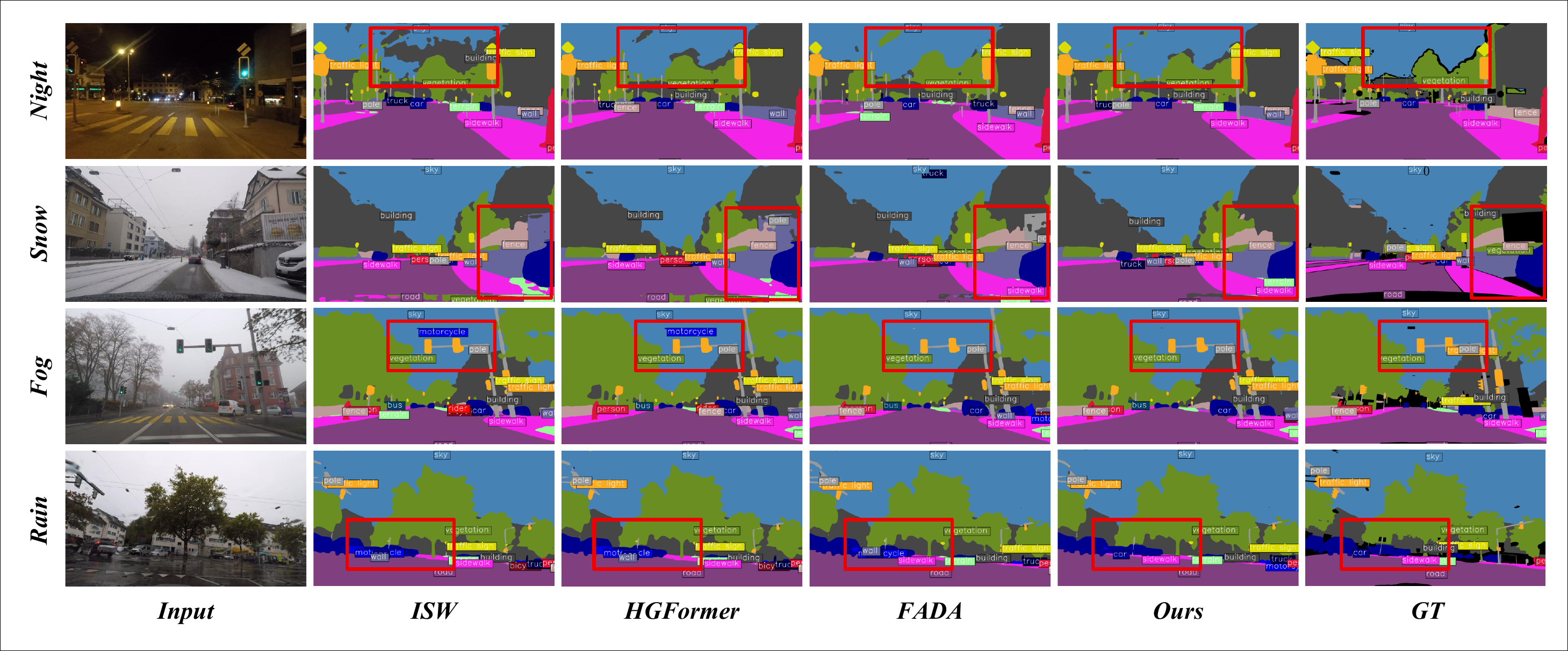}
\caption{Visualization segmentation results under $C.$ $\rightarrow$ ACDC generalization. Causal-Tune is compared with ISW~\cite{ISW} (ResNet-based), HGFormer~\cite{CMFormer} (Transformer-based) and FADA~\cite{FADA} (VFM-based).}
\label{fig_seg_vis_C2ACDC}
\end{figure*}

\subsubsection{Analysis of different frequency domain transformation methods.}
As mentioned earlier, the frequency spectrum derived from the DCT has been shown to better separate causal and non-causal factors compared to methods based on FFT and HWT. In this section, we explore the effects of different frequency domain transformation methods on Causal-Tune. 
From Table~\ref{table_frequency} we can observe that 
HWT cannot reveal causal and non-causal factors well, while FFT is simply better at removing the non-causal factor of ‘Night’. However, our proposed method outperforms FFT and HWT on most of the DGSS setups, which demonstrates that Causal-Tune could mine causal factors and remove non-causal factors better by utilizing DCT.

\subsubsection{Different filtering method on DCT frequency spectrum.}
To further verify the effectiveness of the DCT in our method, we conduct an ablation study on different filtering methods for the DCT frequency spectrum. Specifically, compared to baseline~\cite{rein} (the first row in Table~\ref{table_DCT_component}), when we only remove the low-frequency components (\textit{LF}) in the DCT spectrum, the performance improves under `Night' and `Snow' condition, while it degrades under other conditions (the second row in Table~\ref{table_DCT_component}). Moreover, from the third row of Table~\ref{table_DCT_component}, removing high-frequency (\textit{HF}) could enhance the generalization of the model under `Snow', `Fog' and `Rain' conditions. In our method, we remove both low- and high-frequency components in the DCT spectrum, \ie,  use a band-pass filter, we achieve the best average performance in four conditions (72.0\% in the last row).

Based on this experiment, we can observe that non-causal factors like `Fog' and `Rain' are mostly concentrated in the high-frequency of the DCT spectrum, whereas non-causal factors of `Night' are mostly concentrated in the low-frequency of the DCT spectrum. Moreover, `Snow' exists in both low- and high-frequency and a band-pass filter could remove these non-factors more effectively.

\subsubsection{Analysis of the cutoff frequency parameters for the band-pass filter.}
To verify the impact of the band-pass filter cutoff frequencies in our method on the DGSS task, we conduct experiments under $C. \rightarrow$ ACDC setting with different cutoff frequencies and compute the corresponding accuracy matrix. As shown in Figure~\ref{fig_acc_matrix}, when low-cutoff frequency $R_L\leq$ is 0.2 and high-cutoff frequency $R_L\leq$ is 0.8, the model generalizes better under four adverse weather conditions. However, as both cutoff frequencies increase ($R_L>$ 0.2, $R_H>$ 0.8), the performance of the model degrades. Thus, we set $R_L =$ 0.2, $R_H=$ 0.7 as the default.

\subsection{Visualization Results}

\subsubsection{Feature map visualization of different VFMs.}
As mentioned before, the artifacts in the feature of VFMs may be caused by implicit non-causal factors that are not fully eliminated during the fine-tuning. In Figure~\ref{fig_feature_map}, we observe that the artifacts in EVA02 are significantly reduced after fine-tuning by our method. Moreover, the artifacts in DINOv2 and CLIP are also removed. This demonstrates that removing non-causal factors can reduce artifacts in VFMs and allow models to focus on domain-invariant causal factors.

\subsubsection{Visualization of segmentation results.}
To further show the effectiveness of our Causal-Tune, we present some visualization results on $C. \rightarrow$ ACDC setting. As shown in Figure~\ref{fig_seg_vis_C2ACDC}, 
our method shows better segmentation results not only the ResNet-based method ISW~\cite{ISW} and the Transformer-based method HGFormer~\cite{CMFormer}, but also VFM-based method FADA~\cite{FADA}.

\section{Conclusion}
In this paper, we propose a novel VFM-based PEFT method named Causal-Tune for DGSS from a causal perspective. In order to explicitly mine the causal and non-causal factors of features within VFMs for DGSS, our method incorporates two key components: i) Causal \& Non-causal Factors Filter to disentangle causal factors and non-causal factors within the frequency domain, and ii) Causal Factors Tune, which utilizes a set of causal-aware learnable tokens to refine the causal factors to enhance the ability of generalization. Extensive experiments conducted on various cross-domain tasks show the effectiveness of our method. However, the generalization of the model is somewhat sensitive to the high- and low-cutoff frequency of the band-pass filter, and we plan to design a dynamic cutoff frequency to allow the model to adapt to various conditions in the future.

\section{Acknowledgments}
This work is supported by the National Natural Science Foundation of China (No. 62206077), the Inner Mongolia Natural Science Foundation for Distinguished Young Scholars (No. 2025JQ009), and the Inner Mongolia Talent Development Project for Outstanding Young Talents. Yaoyue
Zheng acknowledges the China Scholarship Council (CSC) No.202406280387. 
This work is supported by Grant PID2022-143257NB-I00 funded by MICIU/AEI/10.13039/501100011033 and ERDF/EU, the Departament de Recerca i Universitats from Generalitat de Catalunya with reference 2021SGR01499, and the Generalitat de Catalunya CERCA Program.

\bibliography{aaai2026}

\end{document}